\documentclass[runningheads]{llncs}
\usepackage[T1]{fontenc}
\usepackage{graphicx}
\usepackage{enumitem}
\usepackage{longtable}
\usepackage{multirow}
\usepackage{pifont}
\newcommand{\cmark}{\ding{51}}%
\newcommand{\xmark}{\ding{55}}%
\usepackage{rotating}
\usepackage{booktabs}
\usepackage[table]{xcolor} 
\usepackage{colortbl}
\usepackage{appendix}
\definecolor{PrivateRedOld}{rgb}{1.0, 0.3412, 0.349}
\definecolor{PublicGreenOld}{rgb}{0.55412, 0.6941, 0.49}
\definecolor{LightGrey}{rgb}{0.95, 0.95, 0.95}
\definecolor{DarkGrey}{rgb}{0.37, 0.37, 0.37}

\definecolor{PrivateOrange}{rgb}{1.0, 0.419608, 0.20784}
\definecolor{PublicBlue}{rgb}{0, 0.305882, 0.53725}
\definecolor{ZeroWeightBrown}{rgb}{0.4313, 0.23137549, 0.160784314}

\definecolor{PrivacyAlertYellow}{rgb}{0.90980392, 0.65098039, 0.42352941}
\definecolor{VISPRBlue}{rgb}{0.19215686, 0.35294118, 0.52941176}
\definecolor{DIPA2Purple}{rgb}{0.52941176, 0.30980392, 0.45882353}

\definecolor{VISPRDescriptionsMint}{rgb}{0.37647059, 0.73333333, 0.71764706}
\definecolor{LegalDescriptionsPink}{rgb}{1.        , 0.73333333, 0.71764706}

\begin{document}

\title{PrivLEX: Detecting legal concepts in images through Vision-Language Models}
\titlerunning{PrivLEX: Detecting legal concepts in images through VLMs}

\author{Darya Baranouskaya\inst{1,2} \and
Andrea Cavallaro\inst{1}}
\authorrunning{D. Baranouskaya and A. Cavallaro}

\institute{EPFL, Lausanne, Switzerland
\and
Idiap Research Institute, Martigny, Switzerland\\
\email{\{darya.baranouskaya,andrea.cavallaro\}@epfl.ch}}
\maketitle              
\begin{abstract}
We present PrivLEX, a novel image privacy classifier that grounds its decisions in legally defined personal data concepts. PrivLEX is the first interpretable privacy classifier aligned with legal concepts that leverages the recognition capabilities of Vision-Language Models (VLMs). PrivLEX relies on zero-shot VLM concept detection to provide interpretable classification through a label-free Concept Bottleneck Model, without requiring explicit concept labels during training. We demonstrate PrivLEX's ability to identify personal data concepts that are present in images. We further analyse the sensitivity of such concepts as perceived by human annotators of image privacy datasets.

\end{abstract}

\section{Introduction}

Sharing photos on digital platforms can inadvertently expose personal information. Identifying these potential leaks with a classifier may help protect privacy and prevent misuse of personal information. However, the subjective nature of privacy perception influences the consistency of annotations in image privacy classification datasets~\cite{Orekondy_2017,Zhao_2022_privacyalert}, thus making the problem of training a classifier a challenging task.

Semantically meaningful concepts that describe image content are commonly used to ground model predictions in a human-understandable way~\cite{Ayci_2023_PEAK,baia2024image}. 
 Vision-Language Models (VLMs) can be used to extract privacy concepts from images in a zero-shot setup~\cite{baia2025_zero_shot_image_privacy,li2025egoprivacy,samson2024little_data,sun2025multipriv,tsaprazlis2025_assessing_visual_privacy_risks,zhang2024multip2a}. VLMs have been benchmarked for privacy awareness~\cite{zhang2024multip2a} and reasoning~\cite{sun2025multipriv}, inference of privacy concepts from first-person view videos~\cite{li2025egoprivacy} and extraction of high-level privacy categories, such as biometric data, financial information and social data~\cite{samson2024little_data,sun2025multipriv,tsaprazlis2025_assessing_visual_privacy_risks,zhang2024multip2a}.
These methods usually analyse only a small number of concepts (e.g.~age, race and gender)~\cite{li2025egoprivacy,samson2024little_data}, broad categories~\cite{sun2025multipriv,tsaprazlis2025_assessing_visual_privacy_risks,zhang2024multip2a} or binary privacy labels~\cite{baia2025_zero_shot_image_privacy}. 

Privacy classifiers have used concepts and textual descriptions extracted from images, such as privacy-specific tags~\cite{8features} and information about the presence and number of people~\cite {xompero2024explaining}. Interpretable privacy classifiers provide explanations through general concepts~\cite{Ayci_2023_PEAK,baia2024image} or require laborious annotation of each privacy-related concept~\cite{Orekondy_2017}.

In this paper, we aim to identify potential personal information leaks that may result from sharing a photo. To this end, we propose an image privacy classifier, PrivLEX, that grounds its predictions in legally defined personal data concepts extracted from data protection documents~\cite{ai_act,gdpr,dga}. PrivLEX builds on label-free Concept Bottleneck Models (CBMs)~\cite{oikarinen2023label_free_concept_bottlenecks,yang2023language_model_guided_bottlenecks_for_interpretable_image_classification,yuksekgonul2022post_hoc_bottleneck}, where the concepts are detected in images with VLMs in a zero-shot manner, without the need for explicit concept labels during training. 
The final classification decision is made as a linear combination of the predicted concept probabilities. We train the final linear classifier on binary subjective human-privacy preferences~\cite{Zhao_2022_privacyalert} to identify which legally defined personal data concepts are perceived as important by humans. Furthermore, we use  PrivLEX to identify personal data concepts that emerge across different privacy datasets and uncover biases in image selection and annotation. The code is available at https://github.com/idiap/privlex/.

\section{Related Work}
\label{sec: related works}

To provide explanations of image privacy through natural language concepts, descriptions extracted from privacy datasets can be used~\cite{Ayci_2023_PEAK,baia2024image}. PEAK~\cite{Ayci_2023_PEAK} groups general tags from images into topics. Priv×ITM~\cite{baia2024image} leverages VLM-generated captions to discover sets of human-understandable descriptors from clusters of images. These general data-driven topics may miss important concepts that are underrepresented in privacy datasets.

AP-PR~\cite{Orekondy_2017} was the first method to utilise a visual privacy taxonomy with a list of 67 privacy concepts, derived from social media privacy policies, legal documents, and manual inspection of images. However, this approach requires images to be labelled with respect to each concept to provide an explanation, which limits the scalability and adaptation to personal data concepts not covered in the corresponding annotated dataset (VISPR). Other visual privacy taxonomies, VisPT~\cite{tsaprazlis2025_assessing_visual_privacy_risks} and MultiPriv~\cite{sun2025multipriv}, were inspired by legal data but focus on a subset of personal data and broad data categories. To the best of our knowledge, PrivLEX is the first method for image privacy classification aligned with a legal definition of privacy.

VLMs compute concept scores in a zero-shot setup~\cite{menon2022visual_classification_via_description_from_llm,oikarinen2023label_free_concept_bottlenecks,yang2023language_model_guided_bottlenecks_for_interpretable_image_classification} for concrete visual concepts, such as objects, in the image~\cite{cooper2025rethinking_vlm_and_llm_for_image_cls,yang2023language_model_guided_bottlenecks_for_interpretable_image_classification}. While VLMs can ground (some) concepts~\cite{cooper2025rethinking_vlm_and_llm_for_image_cls,menon2022visual_classification_via_description_from_llm}, concepts that are not visually discriminative are rarely included in the studies~\cite{zang2024pre_trained_vlms_learn_discovrable_visual_cocnepts}. Examples of these more abstract, less visually discriminative concepts include political opinion, personal relationships and educational history, which are important for subjective tasks like image privacy classification~\cite{Orekondy_2017}. 

Concept Bottleneck Models (CBMs) align model predictions with human-understandable concepts~\cite{koh2020concept_bottleneck_models,oikarinen2023label_free_concept_bottlenecks,yuksekgonul2022post_hoc_bottleneck}. First, a set of concepts with corresponding concept scores is predicted from the image, and then a linear combination of these concepts is used to predict the final class. As traditional CBMs require human annotation of concepts, several studies have leveraged natural language concept descriptions and multi-modal models, like CLIP~\cite{CLIP}, to compute concept scores in a zero-shot manner~\cite{oikarinen2023label_free_concept_bottlenecks,yang2023language_model_guided_bottlenecks_for_interpretable_image_classification,yuksekgonul2022post_hoc_bottleneck}. While such models can accommodate a larger set of concepts, label-free CBMs used for image privacy are not directly grounded in privacy laws~\cite{baia2024image}.

\section{Detecting Legal Concepts in Images}

We adopt a list of legally defined concepts related to personal data as a bottleneck of a label-free CBM. 
We use the concepts and their short description that appear in the Data Privacy Vocabulary (DPV-PD)~\cite{pandit2024dpv_pd2}, which aggregates different types and categories of personal data based on the General Data Protection Regulation (GDPR)~\cite{gdpr}, the Data Governance Act (DGA)~\cite{dga} and the AI Act~\cite{ai_act}. 
DPV-PD defines personal data as data directly or indirectly associated with or related to an individual (Table~\ref{tab: DPV-PD hierarchy}).
\begin{table}[t]
\centering
\caption{DPV-PD terms presented with respect to a four-level hierarchy. } 
\tiny
\begin{tabular}{p{1.3cm}p{2.5cm}p{7.5cm}}
\textbf{First level} & \textbf{Second level} & \textbf{Third (and Forth) level} \\
\toprule
\multirow{3}[1]{*}{Internal} & Authenticating & Password, PIN Code, Secret Text\\
 & Knowledge Belief & Philosophical Belief, Religious Belief, Thought  \\

 & Preference & Favorite (e.g. Favorite Color, Favorite Food), Intention, Interest (Dislike, Like), Opinion, Privacy Preference\\

\midrule
\multirow{12}[1]{*}{External} & Behavioral & e.g. Attitude, Browsing Behavior (e.g. Browser History), Link Clicked, Personality\\
 &  Citizenship &  \\
 &  Demographic &  Geographic, Income Bracket, Physical Trait\\
 &  Ethnicity &  Ethnic Origin, Race\\
 &  Identifying & e.g. Biometric (e.g. Fingerprint), Official ID (Passport), Username\\
 & Language  &  Accent, Dialect\\
 &  Medical Health &  e.g. Blood Type, Disability, Health (e.g. Genetic, Mental Health), Prescription\\
 &  Nationality &  \\
 &  Personal Documents &  \\
 &  Physical Characteristic & e.g. Age (Age Range, Birth Date), Gender, Skin Tone, Tattoo \\
 &  Sexual & Fetish, Proclivity, Sexual History, Sexual Preference \\
 &  Vehicle &  Vehicle License (Vehical License Number, Vehical License Registration), Vehicle Usage\\

 \midrule
\multirow{5}[1]{*}{Financial} & Financial Account & Account Identifier (e.g. Financial Account Number), Bank Account, Payment Card (e.g. Payment Card Expiry)\\
& Financial Status & \\
& Insurance & \\
& Ownership & Car Owned, House Owned (Apartment Owned), Personal Possession\\
& Transactional & e.g. Credit (e.g. Credit Record), Income, Tax\\

\midrule

Historical & Life History\\
\midrule

\multirow{6}[1]{*}{Social} & Communication & e.g. Email Content, Social Media (Publicly Available Social Media), Voice Mail\\
 & Criminal & e.g. Criminal Charge, Criminal Offense\\
 & Family & e.g. Family Structure (e.g. Divorce, Offspring), Relationship\\
 & Professional & e.g. Disciplinary Action, Employment History (Current Employment, Past Employment), Salary \\
 & Public Life & e.g. Character, Political Affiliation, Religion\\
 & Social Network & e.g. Association, Connection, Group Membership (Trade Union Membership)\\

\midrule
\multirow{6}[1]{*}{Tracking} & Contact & Email Address (Email Address Personal, Email Address Work), Physical Address, Telephone Number \\
 & Device Based & e.g. Browser Fingerprint, Device Software (Device Applications, Device Operating System), IP Address\\
 & Digital Fingerprint & \\
 & Identifier & \\
 & Location & Country, GPS Coordinate, Physical Address (e.g. House Number, City, Room Number), Travel History\\
 & User Agent & \\

\bottomrule
\end{tabular}

\label{tab: DPV-PD hierarchy}
\end{table}
At the first hierarchical level, DPV-PD structures personal data into eight broad groups based on origin and relevance. The second level and the third (if present) comprise the majority of the concepts covering personal data. The third level specifies the second-level concepts. The fourth level provides more detailed subdivisions for a small number of concepts. For our purposes, we selected concepts primarily at the third level of granularity, or at the second level if the third was unavailable, to avoid redundancy. We omit the fourth level due to excessive detail; instead, these concepts are included in the description of the corresponding third-level concepts as examples in braces after “e.g.”. For example, we extend the description of the \textit{browsing behavior} (3rd level) as \textit{“Information about browsing behavior (e.g., browser history, browsing referrals).”}, where \textit{browser history} and \textit{referrals} are corresponding 4th level concepts.
As a result, we obtain $n=131$ personal data concepts, each accompanied by a brief description. This aggregation enables us to provide detailed explanations while minimising semantic overlap among concepts\footnote{Note that our classifier performs well even when using all DPV-PD concepts, ignoring the hierarchy, and on a previously released version of DPV-PD, with only slight (0.4 to 1.3~p.p.) changes to the performance.}.

Label-free CBMs use a textual description of a concept to compute concept scores between an image and that concept using a VLM~\cite{oikarinen2023label_free_concept_bottlenecks,yang2023language_model_guided_bottlenecks_for_interpretable_image_classification}. 
As VLM, we chose CLIP~\cite{CLIP} as it is widely used~\cite{cooper2025rethinking_vlm_and_llm_for_image_cls,menon2022visual_classification_via_description_from_llm} and offers acceptable performance with high computational efficiency for the task.
We learn the mapping between the DPV-PD concepts and the binary image class by training a CBM on privacy annotations. 

Given an image $x_i$, the VLM image encoder produces an embedding $I_i \in R^d$, where $d$ is the dimension of the embedding space shared by both images and textual inputs (sentences). For each DPV-PD personal data concept, $t_j$ with $j=1, \dots, n$, the VLM text encoder produces an embedding $T_j \in R^d$ by encoding the corresponding sentence, constructed as: \\\\\texttt{<concept>: <description>}.\\

We quantify the presence of concept $t_j$ in image $x_j$ using a concept score, $c_{ij}$, which is computed as the cosine similarity between the corresponding image embeddings $I_i$ and concept embeddings $T_j$: 
\begin{equation}
  c_{ij} = \frac{I_i \cdot T_j}{||I_i|| ||T_j||},   
\end{equation}
where $\cdot$ denotes the dot product.

The concept scores are normalised to the range [0, 1], yielding  (${c_{i1}^N, ..., c_{in}^N}$), and passed to a Logistic Regression (LR) classifier. The LR model, parameterised by the weights $W=(w_1, ..., w_{n})$ and bias $b$, is trained to predict a binary privacy label on a privacy classification dataset. We optimise the model using binary cross-entropy. To enhance interpretability and reduce the influence of concepts unrelated to visual privacy, we enforce sparsity of the LR through strong L1 regularisation~\cite{oikarinen2023label_free_concept_bottlenecks}.

During training, the LR captures task-specific knowledge by learning the relation between (legal) personal data concepts and (subjective) privacy labels. The sign of the weight signals whether the corresponding concept contributes to the private or public class, while the magnitude of the weight reflects the strength of this contribution.  Positive LR weights indicate that the corresponding personal data concepts, when detected, are associated with the private labels of images in the corresponding dataset. Negative weights correspond to concepts that, although legally defined as private, do not contribute to the private class when detected. Zero weights correspond to concepts that are not presented in the dataset or are non-discriminative for privacy, either because they carry little privacy-relevant information or are highly correlated with other concepts assigned non-zero weights.

Since the model tends to assign larger magnitudes to concepts it detects more consistently, the magnitude of LR weights may be affected by the uneven distribution of content (and mislabels\footnote{For example, private concepts like \textit{fingerprint} or \textit{payment card} are present in a few images only, therefore rare mislabels and VLM recognition errors will significantly affect the concept's probability of contributing to the private class.}) in the training dataset~\cite{king2001lr_in_rare_events_data}. To account for this, we adopt a dual strategy for the analysis. We focus on {\em both} the sign and magnitude of the LR weights learned for each dataset when we investigate biases of privacy datasets towards particular types of content or patterns shared across datasets.
In contrast, when presenting personal data concepts detected in an image in order to minimise the influence of training data scarcity, we focus {\em solely on the sign} of the weights, disregarding (unlike in standard CBMs) their magnitude.

The personal data concepts with the highest concept scores are most likely present in an image. To reveal information about personal data concepts in an image $x_i$ to users, we output the $k$ concepts with the highest concept scores, 
\begin{equation}
    k = max\left(\sum_{j=1}^n\textbf{I}_{[c_{i, j} > \tau]}, 3\right),
\end{equation}
along with the sign of their corresponding LR weights. We present all concepts with scores above an empirically determined threshold $\tau$, and at least three concepts per image.

\begin{figure}[t]
    \centering
    \includegraphics[width=1\textwidth]{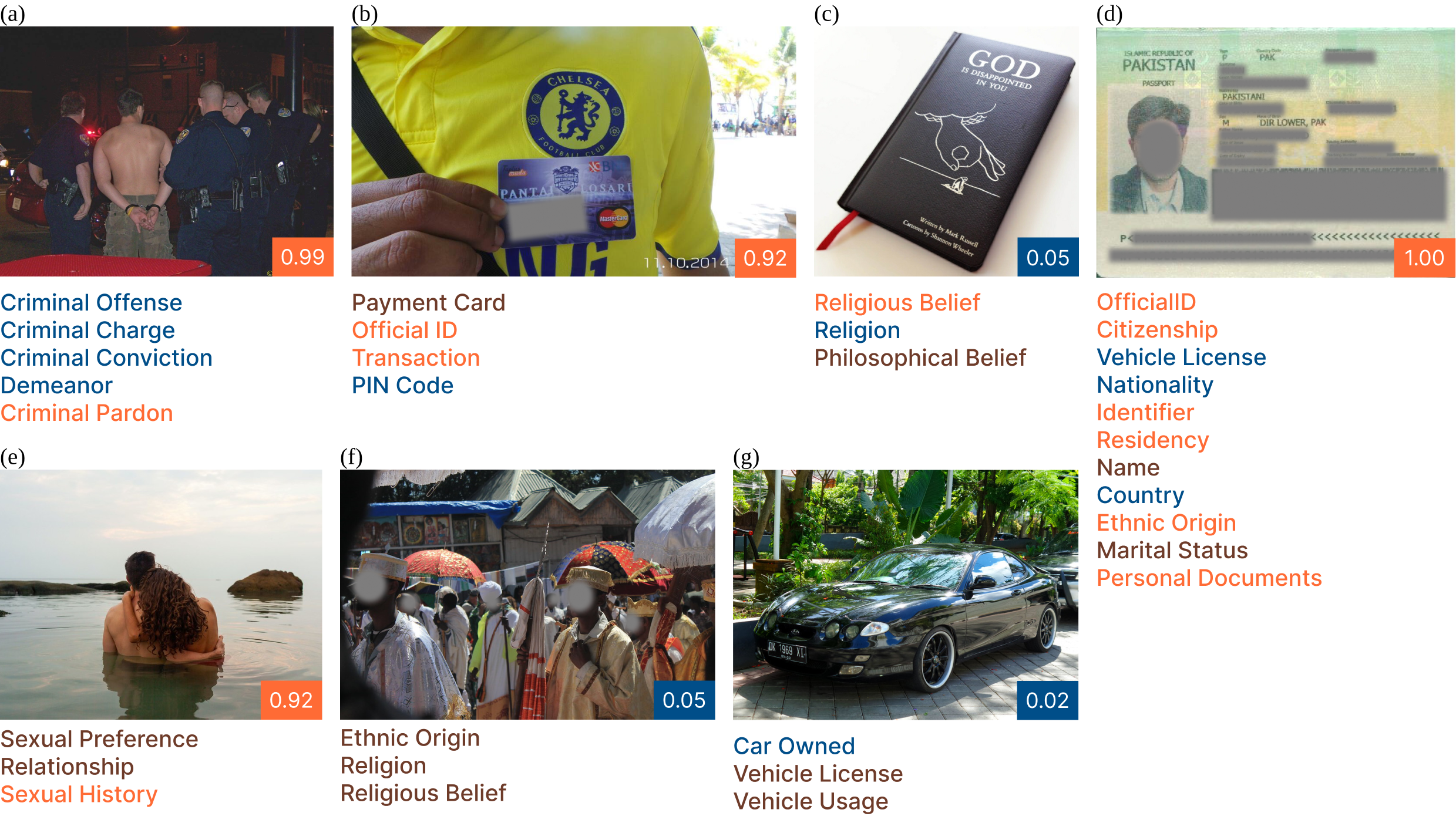}
    \caption{Identified personal data concepts by PrivLEX for VISPR~\cite{Orekondy_2017} (a-d) and PrivacyAlert~\cite{Zhao_2022_privacyalert} (e-g) images. The colour in the bottom-right corner of an image indicates the ground-truth label (\textcolor{PrivateOrange}{private}/\textcolor{PublicBlue}{public}), and the number corresponds to the predicted probability of the private class. The colour of a concept indicates its contribution to the \textcolor{PrivateOrange}{private} or \textcolor{PublicBlue}{public} class, or signals \textcolor{ZeroWeightBrown}{zero weight}.}
    \label{examples of correctly classified VISPR images}
\end{figure}

Fig.~\ref{examples of correctly classified VISPR images} shows examples of personal data concepts identified by PrivLEX.

\section{Experiments}
\subsection{Experimental setup}

We compare PrivLEX with the four most recent interpretable privacy classifiers, namely PEAK~\cite{Ayci_2023_PEAK}, Priv×ITM~\cite{baia2024image}, Strategy-2~\cite{xompero2024explaining}, LR on 8PS~\cite{8features}, and two non-interpretable models with the best performance on PrivacyAlert, namely  SVMxIB~\cite{baia2024image} and GATED~\cite{Zhao_2022_privacyalert}. 
We use two image privacy datasets, namely PrivacyAlert~\cite{Zhao_2022_privacyalert} and VISPR~\cite{Orekondy_2017}. VISPR was annotated with respect to 67 privacy-related concepts or {\em safe} if no concept is present. To receive binary (private/public) labels for images in VISPR, we binarised the annotation of the {\em safe} attribute.
Due to class imbalance in the datasets, as performance measures, we consider balanced accuracy (BA) and F1-macro score (F1-m) as primary evaluation measures. We also report the F1-score for the private class to control the balance between minimising private information leakage and avoiding overly restrictive predictions.

To choose the value of the {\em hyperparameters} when training LR in PrivLEX, we optimise the F1-macro on the validation set for each dataset with Optuna\footnote{https://optuna.org} on 100 steps across the following grid values: $C$: from $1 \times e^{-10}$ to 1 to promote sparsity, $max\_iter$: from 1 to 250. We fix the hyperparameters for the ablation of privacy-related concepts. However, we observe that the qualitative conclusions of the ablation remain unchanged when hyperparameters are instead optimised independently for each concept type. Based on the analysis of concept scores for relevant and irrelevant concepts on a random set of images from the training set, we choose $\tau=0.245$.

\subsection{Results and discussion}

\noindent{\bf Classification results}. Table~\ref{tab: Comparison with other methods} shows that PrivLEX outperforms state-of-the-art interpretable classifiers in terms of  BA and F1-m on both datasets. We report the average performance of PrivLEX across 10 runs. As the standard deviation consistently remains below 0.1\%, we do not report it in the table. PrivLEX outperforms the previous state-of-the-art method, Priv×ITM, by 1.02 percentage points (p.p.) and 0.45~p.p. in both BA and F1-m on PrivacyAlert, and by 1.78~p.p. and 1.78~p.p. on VISPR. PrivLEX also performs  better than other interpretable models, PEAK, Strategy-2 and LR on 8PS.
Finally, as a reference, we evaluate the GPT-4o model\footnote{https://platform.openai.com/docs/models/gpt-4o} on PrivacyAlert. To obtain the GPT-4o privacy predictions, we pass an image and the prompt used to elicit the annotators' responses as input and request, as output, a binary label (private/public). PrivLEX outperforms GPT-4o in both F1-m and BA. Additionally, PrivLEX matches the performance of the non-interpretable highest-performing classifiers SVMxIB and GATED in BA, while they surpass PrivLEX by 2.49 p.p and 0.65~p.p in F1-m. The lower F1-m is attributable to lower precision on the private class, likely due to the VLM overpredicting concepts. 
\begin{table}[t!]
\caption{Comparison to existing classifiers: three best-performing non-interpretable and four interpretable methods, primarily focusing on the newer PrivacyAlert~\cite{Zhao_2022_privacyalert} dataset. The most relevant concept-based interpretable classifiers for comparison are Priv×ITM~\cite{baia2024image} and PEAK~\cite{Ayci_2023_PEAK}. For a detailed comparison in addition to the main measures, BA and F1-m, we report accuracy, and precision and recall for each class.}
\begin{tabular}{ccc>{\columncolor{LightGrey}}c>{\columncolor{LightGrey}}cccccccc}
\toprule
& &  \multicolumn{3}{c}{\textbf{Overall}} & \multicolumn{3}{c}{\textbf{Private}} &  \multicolumn{3}{c}{\textbf{Public}}\\
\textbf{DS}  & \textbf{Model} & \textbf{ACC}&  \textbf{BA} & \textbf{F1-m} & \textbf{P} & \textbf{R} & \textbf{F1}  &\textbf{P} & \textbf{R} & \textbf{F1}& \textbf{IP} \\

\midrule
\multirow{7}[1]{*}{\begin{sideways}PrivacyAlert \end{sideways}}& SVMxIB~\cite{baia2024image} & 89.11 &  83.19 & 85.39 & 78.73 & 73.33 & 78.03 & 92.49 & 93.04 & 92.76 & \multirow{3}[1]{*}{\xmark} \\
 
 & GATED~\cite{Zhao_2022_privacyalert}  & 87.94 & 82.70 & 83.55  & 77.90 &  72.20 & 75.00  & 91.00 &  93.20 &  92.10 &  \\

& GPT-4o\footnote{https://platform.openai.com/docs/models/gpt-4o} & 75.90 & 77.71 & 72.50 & 51.19 & 81.33 & 62.83 & 92.24 & 74.09 & 82.17 &  \\

\cmidrule{2-12}

& LR on 8PS~\cite{8features} & 80.67  & 80.44  & 76.84  & 58.25  & 80.00  & 67.42  & 92.39  & 80.89  & 86.26  & \multirow{10}{*}{\cmark} \\

& Strategy-2~\cite{xompero2024explaining} & 73.83 & 74.11 &  69.83 &  48.55 & 74.67  & 58.84 & 89.67 & 73.55 & 80.81 & \\

& PEAK~\cite{Ayci_2023_PEAK} & 84.61 & 76.56  &  78.15 &  73.32  &  60.44 & 66.26 & 87.54 &  92.67 & 90.00   &  \\

& Priv×ITM~\cite{baia2024image} & 86.94 &  82.19 & 82.45 & 74.49 & 72.67 & 73.57 & 90.96 & 91.70 & 91.33 &  \\

& PrivLEX & \textbf{87.05}  & \textbf{83.21} & \textbf{82.90} & 73.43 & 75.53 & 74.47 & 91.77 & 90.89  & 91.33\\

\cmidrule{1-11}

\multirow{3}{*}{\begin{sideways}VISPR\end{sideways}} & PEAK$^{*}$~\cite{Ayci_2023_PEAK} & 85.47 & 83.58  &  84.20 &  86.30  &  91.20 & 88.68  & 83.86  &  75.96 & 79.71   &  \\

& Priv×ITM~\cite{baia2024image} & 87.16 & 86.33 & 86.32 & 89.74 & 89.69 & 89.71  & 82.88 & 82.97 & 82.93  \\

& PrivLEX & \textbf{88.84} & \textbf{88.11} & \textbf{88.10} & 91.07 & 91.05 & 91.06 & 85.14 & 85.16 & 85.15 \\

\bottomrule
\addlinespace[\belowrulesep]
\multicolumn{12}{l}{\parbox{0.9\linewidth}{\scriptsize{
Key -- DS: dataset, IP: signals if the method is interpretable, ACC: accuracy, BA: balanced accuracy, F1-m: F1-macro score, F1: F1 score, P: precision, R: recall, $^{*}$: the test set contains 7970 images instead of the standard 8000. 
}}}

\end{tabular}

\label{tab: Comparison with other methods}
\end{table}

\noindent{\bf CLIP recognition capabilities}.
\label{sec: discussion, vlm recognition}
To assess the quality of CLIP zero-shot recognition of legal concepts, we use the VISPR annotated concepts~\cite{Orekondy_2017} as a reference, since the DPV-PD concepts are not annotated in images. We manually generate legal-style descriptions for the VISPR concepts, similar to those DPV-PD have. These descriptions are either derived from similar DPV-PD concepts or created using the template \\ \\ \texttt{<concept>: information about <concept>}. \\ \\
\begin{figure*}[t!]
    \centering
    \includegraphics[ width=0.42\textwidth]{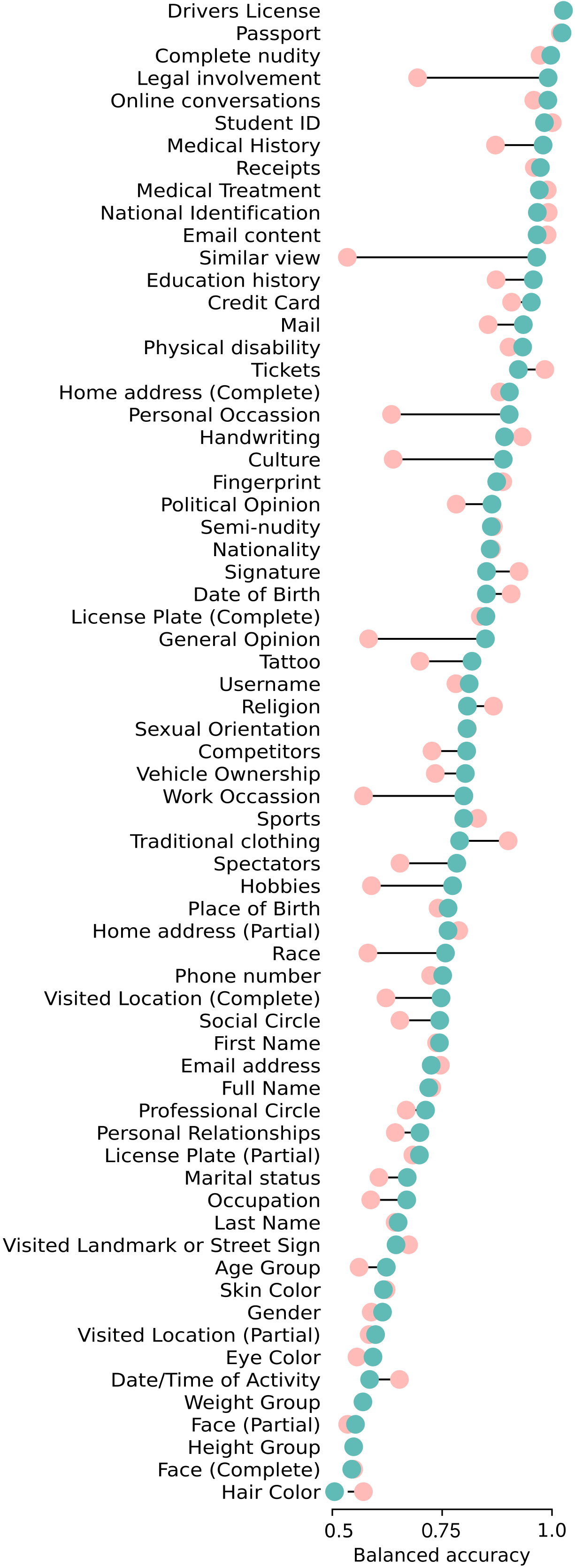}
    \caption{
    Balanced accuracy of CLIP zero-shot recognition of VISPR concepts~\cite{Orekondy_2017} with two types of concept descriptions: 1) detailed VISPR descriptions provided in the dataset for annotators (VISPR-VI) {\textcolor{VISPRDescriptionsMint}{\large$\bullet$}}, 2) legal-style descriptions (VISPR-LS) {\textcolor{LegalDescriptionsPink}{\large$\bullet$}}.}
    \label{img: CLIP zero-shot recognition of VISPR concepts with VISPR and legal-like descriptions}
\end{figure*} 
Concept detection is performed by thresholding the concept score produced by CLIP for an image and the textual description of a concept. We select the thresholds for each concept based on the best balanced accuracy achieved on the training set.
\begin{figure}[t]
    \centering
    \includegraphics[width=1\textwidth]{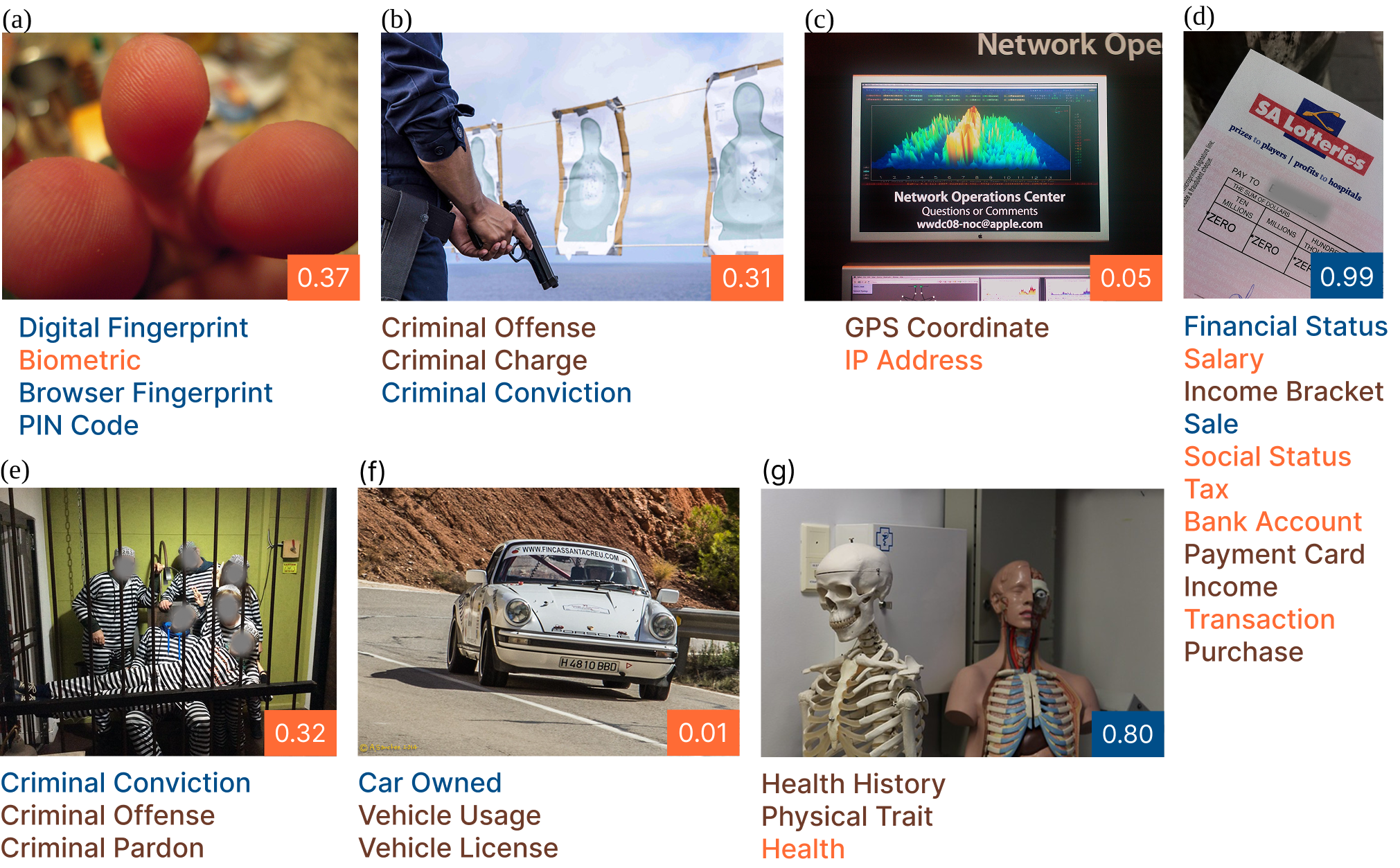}
    \caption{Misclassified by PrivLEX images from VISPR~\cite{Orekondy_2017} (a-d) and PrivacyAlert~\cite{Zhao_2022_privacyalert} (e-g) with identified personal data concepts. The colour in the bottom-right corner of an image indicates the ground-truth label (\textcolor{PrivateOrange}{private}/\textcolor{PublicBlue}{public}), and the number corresponds to the predicted probability of the private class. The colour of a concept indicates its contribution to the \textcolor{PrivateOrange}{private} or \textcolor{PublicBlue}{public} class, or signals \textcolor{ZeroWeightBrown}{zero weight}.}
    \label{Misclassified_VISPR_and_PrivacyAlert}
\end{figure}

Fig.~\ref{img: CLIP zero-shot recognition of VISPR concepts with VISPR and legal-like descriptions} shows the balanced accuracy of zero-shot CLIP recognition of VISPR concepts with two types of descriptions: legal-style (VISPR-LS) and the descriptions provided for the annotators of the VISPR dataset (VISPR-VI).
On the one hand, CLIP can recognise object-related as well as less concrete concepts that require contextual information. On the other hand, concepts with a wide range of possible visual representations and textual concepts tend to have lower detection accuracy. 

The average balanced accuracy achieved by CLIP across all concepts is above 72.12~p.p. for both description types. 
The median difference between CLIP's BA on VISPR-VI and VISPR-LS is only 1.80~p.p., meaning that for the majority of the concepts, legal-style descriptions are nearly as informative as the initial VISPR descriptions. CLIP accurately detects concrete or object-related concepts, such as \textit{passport}, \textit{signature}, and \textit{physical disability}. Moreover, it detects with both description types some of the more abstract concepts that require a good understanding of visual content, such as \textit{medical treatment}, \textit{education history}, \textit{nationality}, and \textit{religion}. 
However, several concepts, such as \textit{legal involvement}, \textit{similar view}, and \textit{personal occasion}, exhibit a notable drop when using legal-style descriptions, despite being among the best-recognised with VISPR descriptions. For most such concepts, the drop is explained by the specificity of the annotation details for the VISPR dataset, e.g. for \textit{similar view} VISPR descriptions specify that placards or clothing denoting a cause or rallying should be visible. Such detailed annotation also commonly leads to low precision on legal-style descriptions, with an average of 14.17~p.p. across all concepts. 
Moreover, CLIP has lower accuracy for more abstract or generic concepts like \textit{place of birth}, \textit{weight group}, \textit{personal relationships} and \textit{hobbies}. These concepts correspond to diverse visual cues (e.g.~a variety of objects may signal a hobby), making it difficult for CLIP to distinguish between their presence and weak visual associations.
Similarly, the lack of context in legal descriptions of DPV-PD can lead to recognition errors and misinterpretations, such as mapping of \textit{residency} to medical residency and \textit{digital fingerprint} to physical fingerprint (Fig.~\ref{Misclassified_VISPR_and_PrivacyAlert}a).
As CLIP is not explicitly trained to comprehend text embedded in images, different types of textual personal data, such as \textit{name}, \textit{email address}, \textit{telephone number}, and \textit{PIN code}, may be confused (Fig.~\ref{examples of correctly classified VISPR images}d, Fig.~\ref{Misclassified_VISPR_and_PrivacyAlert}c-d). However, such incorrectly inferred concepts typically receive lower confidence scores, indicating that CLIP can relate embedded text, as shown in  Fig.~\ref{examples of correctly classified VISPR images}c, where \textit{religion} is inferred from the word depicted on the cover of the book. 

\noindent{\bf Alternative privacy-related concepts}. We compare the DPV-PD concepts used in PrivLEX with alternative visual privacy taxonomies, namely   VISPR-VI~\cite{Orekondy_2017} and the recently proposed VisPT~\cite{tsaprazlis2025_assessing_visual_privacy_risks} and MultiPriv~\cite{sun2025multipriv}. 
VISPR-VI contains 67 concepts with corresponding descriptions provided for the VISPR dataset annotation~\cite{Orekondy_2017}. To encode the 14 VisPT categories, which are provided with descriptions and examples, we construct sentences similarly to DPV-PD: \\ \\ \texttt{<concept>: <description>, e.g. <examples>}. \\ \\ The sentences for the 36 MultiPriv concepts, since no descriptions were provided, are constructed as \\ \\ \texttt{<concept>: information about <concept>}. \\ \\  
We also consider personal data concepts generated by GPT-4o (LLM-gen). We include the LLM generated concepts in the comparison since CBMs built on such concepts achieve high performance in other image classification tasks~\cite{menon2022visual_classification_via_description_from_llm,yang2023language_model_guided_bottlenecks_for_interpretable_image_classification}. The prompt used for the concept generation\footnote{The number of requested concepts, 131, equals the number of concepts in DPV-PD.} was:\\ \\{\small \texttt{Generate a list of 131 concepts that define personal data categories, each\\ with a very short description. Personal data is defined as data directly or\\ indirectly associated or related to an individual. The list should cover\\
various types of personal data and include a diverse range of categories.  
\\Return the result in JSON format with each term and its description.}} \\ \\
This process generated 124 personal data concepts. It is important to note that LLM-gen concepts are impacted by prompting strategies\footnote{When the number of requested personal data concepts, 131, is not forced in the prompt, LLM outputs under 10 concepts.} and are not (explicitly) related to privacy laws. 

To compare the influence of the selection of privacy-related concepts on the classification performance, we preserve the main PrivLEX architecture and training strategy while replacing the input concepts in the CLIP text encoder with VISPR-VI, VisPT, MultiPriv and LLM-gen concepts. Specifically, we train an LR on the [0, 1]-normalised cosine similarities between a CLIP image embedding and CLIP embeddings of each concept type.
Table~\ref{tab: DPV abblation} shows the performance of the models on five concept types.
Even though VISPR-VI concepts have more detailed descriptions and were specifically developed for the dataset annotation, DPV-PD concepts outperform VISPR-VI on the VISPR dataset with BA and F1-m 2.78~p.p. and 2.41~p.p. higher. PrivLEX outperforms the CBMs using the recent visual privacy taxonomies, VisPT and MultiPriv, by over 1.88~p.p. in both BA and F1-m on PrivacyAlert and 9.47~p.p. on VISPR.  
\begin{table}[t!]
\caption{Comparison of PrivLEX utilising DPV-PD concepts with visual privacy taxonomies VISPR-VI~\cite{Orekondy_2017}, VisPT~\cite{tsaprazlis2025_assessing_visual_privacy_risks} and MultiPriv~\cite{sun2025multipriv}, and LLM-generated personal data
concepts (LLM-gen). The model using DPV-PD shows the best performance across privacy taxonomies and matches LLM-gen performance.}
\centering
\begin{tabular}{l>{\columncolor{LightGrey}}c>{\columncolor{LightGrey}}cc>{\columncolor{LightGrey}}c>{\columncolor{LightGrey}}cc>{\columncolor{LightGrey}}c>{\columncolor{LightGrey}}cc}
\toprule
 & \multicolumn{3}{c}{\textbf{PrivacyAlert}} &  \multicolumn{3}{c}{\textbf{VISPR}} \\
\textbf{Concept types} & \textbf{BA} & \textbf{F1-m}  & \textbf{F1} & \textbf{BA} & \textbf{F1-m}  & \textbf{F1} \\
\midrule
VISPR-VI & 82.99 & 82.66 & 74.12 & 85.33 & 85.39 & 89.06 \\
VisPT &  80.41 & 81.02 & 71.25 & 77.99 & 78.37 & 84.31 \\
MultiPriv & 78.04 & 78.60 & 67.58 & 78.13 & 78.63 & 84.67 \\
LLM-gen &  \textbf{83.37} &  \textbf{83.34} & \textbf{75.03} & 87.97 & 88.02 & 91.03 \\
DPV-PD & 83.21 & 82.90 & 74.47 &  \textbf{88.11} & \textbf{88.10} & \textbf{91.06} \\
\bottomrule
\addlinespace[\belowrulesep]
\multicolumn{7}{l}{\parbox{0.6\linewidth}{\scriptsize{
Key -- BA: balanced accuracy, F1-m: F1-macro score, F1: F1 score for private class.}}}
\end{tabular}
\label{tab: DPV abblation}
\end{table}
Finally, the performance differences between DPV-PD and LLM-gen are smaller than 0.44~p.p. in BA and F1-m on PrivacyAlert, and 0.14~p.p. on VISPR.

\begin{figure}[h!]
    \centering
    \includegraphics[width=0.4\textwidth]{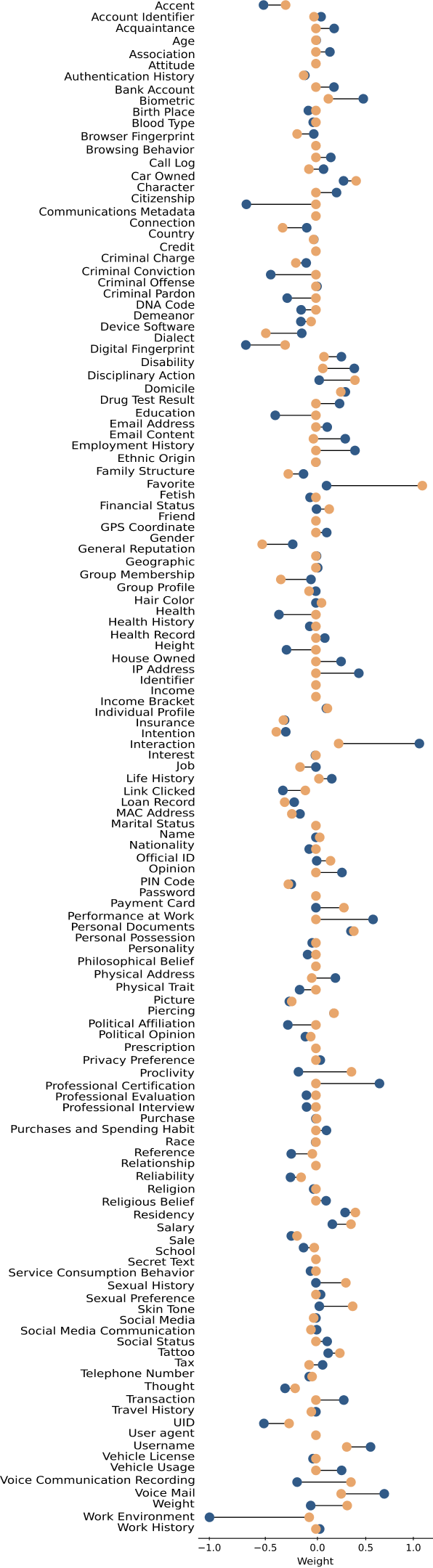}
    \caption{Contribution of legal concepts to the private class (PrivacyAlert{\textcolor{PrivacyAlertYellow}{\large$\bullet$}}, VISPR{\textcolor{VISPRBlue}{\large$\bullet$}}).}
    \label{fig: pra-dipa-vispr_logreg-l1score-Cbelow1-norm[0,1]acrossDatasets PrivLEX LR weights}
\end{figure}
\subsection{Uncovering Dataset Bias(es)}
We use PrivLEX to analyse image privacy datasets as the trained LR weights, $W=(w_1, ..., w_{n})$, reflect the information learned from the dataset. 
For a fair analysis across datasets, we scale the weights of each model to their maximum magnitude and compare the magnitudes of the weights corresponding to different concepts across datasets. PrivacyAlert and VISPR differ in the representativeness of personal data concepts: PrivacyAlert is biased toward content depicting nudity whereas VISPR considers a wide range of concepts related to physical objects and images with people are labeled as private.
By analysing PrivacyAlert and VISPR (see Fig.~\ref{fig: pra-dipa-vispr_logreg-l1score-Cbelow1-norm[0,1]acrossDatasets PrivLEX LR weights}), we found that:
\begin{itemize}[leftmargin=*, labelindent=0pt]
    
    \item Information related to the \textit{work environment}, general characteristics (\textit{general reputation}, \textit{favourite}), or some online tracking concepts (\textit{digital fingerprint}, \textit{links clicked} or \textit{authentication history}) is considered public across datasets. 
    \item Images containing \textit{biometric} and health-related information (\textit{disability}, \textit{drug test result}), \textit{personal documents} or IDs (\textit{professional certification}, \textit{official ID}), names (\textit{username}), and information about sexual preferences (\textit{fetish}) are likely to be labelled as private. 
    \item Concepts related to \textit{work environment}, \textit{performance at work}, vehicle ownership (\textit{car owned}, \textit{vehicle usage}) and sexual life (\textit{fetish}, \textit{proclivity}) are differently represented and evaluated in the two datasets. Images revealing work settings are frequent in VISPR and are commonly public if people are not present and private otherwise, while in PrivacyAlert, work-related concepts have a small contribution. In contrast, information about sexual life is highly private in PrivacyAlert, while in VISPR images, such concepts are rarely present.
    Vehicle-related concepts primarily contribute to the public class in PrivacyAlert and to the private class in VISPR, highlighting annotation contradictions.
\end{itemize}
In summary, the two datasets align in the private perception of biometric and some health-related information, personal IDs and sexual preferences as private, and general personality traits as public.  

\section{Conclusion}
We introduced PrivLEX, a privacy classifier grounded in legally defined personal data concepts. We identified privacy patterns and biases by estimating the perceived sensitivity of these legal concepts using PrivLEX on two image privacy datasets.  PrivLEX achieves state-of-the-art performance across interpretable privacy classifiers and also allows for concept extension or replacement. Moreover, we analysed the possibilities and limitations of recognising legal concepts in images with CLIP VLM. We found that while it may struggle with abstract and vaguely described legal concepts, it generally achieves good recall and rarely misses legal concepts.

Future work includes exploring the use of the DPV-PD four-level hierarchy to enhance child-level concept detection through parent-level categories, implementing a more rigorous cross-validation for the concept detection threshold, and integrating an optical character recognition module to improve document-understanding capabilities.

\subsubsection{Acknowledgements} 
The authors would like to thank Alina Elena Baia for providing the results of the Priv×ITM model on the VISPR dataset. 

\bibliographystyle{splncs04}
\bibliography{main_updated}

\end{document}